\documentclass{article}
\usepackage{ijcai26}

\usepackage{times}
\usepackage{url}
\usepackage[hidelinks]{hyperref}
\usepackage[utf8]{inputenc}
\usepackage[small]{caption}
\usepackage{graphicx}
\usepackage{amsmath}
\usepackage{amsthm}
\usepackage{booktabs}
\usepackage{multirow}
\usepackage[switch]{lineno}
\usepackage{comment}
\usepackage{amssymb}
\usepackage{placeins}
\usepackage{fvextra}
\newcommand{\baseline}[0]{\textsc{Baseline}}
\newcommand{\matrixv}[0]{\textsc{Matrix}}
\newcommand{\mixed}[0]{\textsc{Mixed}}

\title{Clueing up LLMs with Tool-Augmented Deductive Reasoning}

\author{
Rebecca Ansell$^1$
\And
Autumn Toney-Wails$^2,^3$\\
\affiliations
$^1$ Georgetown University\\
$^2$ Syntheos, Corp\\
$^3$ UNU-MERIT\\
\emails
rja80@georgetown.edu,
toney@merit.unu.edu
}

\begin{document}

\maketitle

\begin{abstract}
Despite recent advances in large language models (LLMs), performing logically consistent deductive reasoning over extended interactions remains challenging. Tasks that require integrating evidence across multiple reasoning steps, maintaining consistency with prior inferences, and updating beliefs under new constraints can surface limitations in current models while providing a useful testbed for evaluating reasoning enhancements. In this paper, we implement a text-based, multi-agent version of the classic board game Clue as an environment to evaluate multi-step, agentic deductive reasoning. In this setting, agents must infer hidden information from a sequence of observations, maintain consistency across turns, and reason over an evolving set of logical constraints. We instantiate six LLM-based agents (GPT-4o-mini and Gemini-2.5-Flash) as players that engage in turn-based gameplay; using three agents per model family, we establish baseline performance across repeated games. We then introduce a tool-augmented approach in which a structured possibility matrix converts implicit game state from generated reasoning logs into an explicit representation of remaining possibilities. The possibility matrix encodes extended-turn memory and deductive constraints, offloading these tasks from the agent. We compare this approach against the baseline to evaluate how tool augmentation supports reasoning quality and task success for autonomous agents in a strategic reasoning environment.  

\end{abstract}

\section{Introduction}

Game environments have proven to be dynamic testbeds for researchers to evaluate autonomous agents' abilities to reason and strategize, as these environments extend beyond static benchmarks and evolve with the actions and behaviors of multiple players \cite{huang2025competing}. Specifically, analyzing generated reasoning outputs of large language model (LLM)-based agents throughout gameplay provides insights into their decision-making processes, strategic planning, and interaction dynamics  \cite{chalamalasetti2023clembench,lin-etal-2025-gamebot,gevers2025you,hu2025gamearena}. However, LLM-based agents often underperform in game environments requiring extended memory and multi-step reasoning, motivating the development of prompt engineering, fine-tuning, and tool-augmentation techniques to support agentic reasoning \cite{trencsenyi2025approximating}.

Recent work has highlighted tool augmentation as a promising approach to improving the reasoning capabilities of LLMs \cite{chen2023chatcot,shim2025tooldial}. 
Tool-augmented language models (TALMs) provide LLMs with access to external tools that can be invoked via APIs and support information retrieval, reasoning, and decision-making \cite{parisi2022talm,qu2025tool}. In this way, tool augmentation (rather than prompt engineering and fine-tuning) offloads complex retrieval and reasoning tasks from the agent’s internal representations to external tool interactions, improving performance on benchmarks \cite{hao2023toolkengpt,das2024mathsensei,ma2024sciagent}. Despite these advantages, effectively incorporating tools into LLM-based agents remains challenging, since tool-use capabilities are not uniformly supported across models and agents often struggle to reliably determine when and how to invoke tools in complex, interactive settings.

\begin{figure}
    \centering
    \includegraphics[width=\columnwidth]{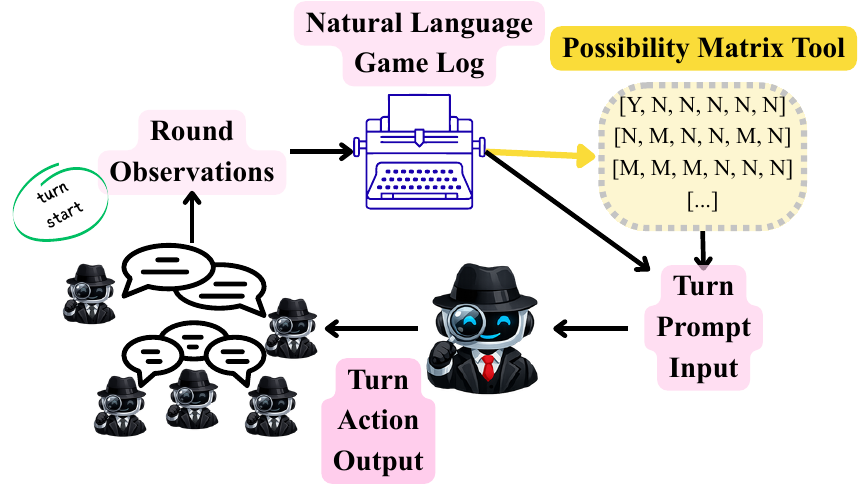}
    \caption{Example turn in Clue game environment with the possibility matrix tool highlighted; the matrix records Yes, No, and Maybe values for possible cards in other players' hands.}
    \label{fig:diagram}
\end{figure}

In this work, we design and implement a tool augmentation approach tailored to interactive game environments by introducing a structured possibility matrix that represents the current game state (shown in Figure \ref{fig:diagram}). Our external tool programmatically extracts and maintains each player’s belief state throughout gameplay, and is invoked at each turn to dynamically update the agent's prompt. By externalizing state tracking in this way, we offload the burden of extended-memory reasoning from the agent and instead provide a consistent and explicit representation of the evolving game state and deductive constraints. Our proposed approach mitigates the degradation of reasoning quality that can arise from long reasoning histories, enabling agents to operate over a cleaner and more reliable contextual grounding when making decisions. Furthermore, our approach is applicable to LLMs that do not natively support tool-use functionalities.

We implement a text-based, multi-agent version of the classic board game Clue following Ansell and Toney~\shortcite{ansell2026how}, which instantiates six agentic players derived from GPT-4o-mini and Gemini-2.5-Flash (three players per model family). Clue provides a deductive reasoning game environment, where game success is achieved through maintaining and updating beliefs over partially observed information to correctly infer the hidden combination of suspect, weapon, and location. To evaluate our tool augmentation approach, we run six games with the baseline models, six games with the players all using the possibility matrix tool, and six games where there are two baseline players and one tool user. 

Across these 18 game instances, we investigate three research questions:

\begin{itemize}
    \item \textbf{(RQ1)} To what extent does externalized belief state tracking via tool augmentation improve agent reasoning and gameplay performance?
    \item \textbf{(RQ2)} How does the presence of tool-augmented agents influence the behavior and performance of non-tool agents in mixed-agent environments? 
    \item \textbf{(RQ3)} How does structured reasoning support affect agents’ decision-making strategies, particularly with respect to risk-taking and the timing of final decisions?
\end{itemize}

We find that tool augmentation leads to near-perfect gameplay for tool-augmented agents, improves the performance of non-tool agents in mixed settings, and results in less delay for accusations without a reduction in accusation accuracy.

\section{Related Work}

\subsection{Deductive Reasoning in LLMs}

Chain-of-thought prompting has achieved measurable improvements on arithmetic, commonsense QA, and symbolic reasoning tasks \cite{chu-etal-2024-navigate}, however these gains degrade as inference chains grow longer, with the generated reasoning steps themselves often containing logical inconsistencies \cite{patel2024multilogievalevaluatingmultisteplogical}. Recent analyses further show that models can arrive at the correct answers through unfaithful intermediate steps, so the surface reasoning does not reliably reflect the process that produced the final answer \cite{xu2026correctnessrewardingfaithfulreasoning,zheng2025cursecotlimitationschainofthought}. Widely used reasoning benchmarks, covering mathematical problem solving \cite{zeng2024mrgsm8kmetareasoningbenchmarklarge,hendrycks2021measuringmathematicalproblemsolving}, commonsense inference \cite{talmor-etal-2019-commonsenseqa}, and formal logic \cite{10174688,yu2020reclorreadingcomprehensiondataset}, share a structural limitation relevant to our setting: events are evaluated in isolation, so they cannot measure whether a model sustains a coherent belief state across multiple interdependent inferences.

\subsection{Games as Deductive Reasoning Testbeds}

Interactive games have become a common way to probe agentic reasoning, but existing environments rarely isolate deduction as the target capability. Social deduction games like Werewolf, Avalon, and Among Us \cite{xu2025languageagentsreinforcementlearning,chi2024amongagentsevaluatinglargelanguage} test persuasion and intent modeling, while Diplomacy \cite{doi:10.1126/science.ade9097}, poker \cite{zhuang2025pokerbenchtraininglargelanguage}, and broader game suites \cite{lin-etal-2025-gamebot} test reasoning under uncertainty. This research gap leaves extended-memory deductive reasoning as an open challenge in game environments. Ansell and Toney~\shortcite{ansell2026how} take a step toward addressing this gap with a text-based implementation of Clue, but finds that models continue to struggle in this setting, even with text-based fine-tuning on related logic puzzle tasks.

\subsection{Tool-Augmented Reasoning}

A growing research area is in Tool-Augmented Language Models (TALMs), which provides LLMs with external tools via API calls \cite{parisi2022talm}. This tool-augmented capability  shifts components of the information retrieval and reasoning workload from the model's internal representations to structured external retrieval and computation \cite{qu2025tool}. Tools are traditionally invoked through LLM API functionality, but can also be encoded in natural language as shown by \cite{hao2023toolkengpt}. TALMs have been shown to improve performance across a range of complex reasoning tasks \cite{chen2023chatcot,das2024mathsensei,ma2024sciagent} as well as in multi-turn, interactive dialogue settings \cite{arcadinho2024automated,shim2025tooldial,jung2025diatool}. However, effective tool use remains an open challenge, as models do not always invoke tools appropriately or reliably integrate tool outputs into downstream reasoning  \cite{patil2024gorilla,chen2024t,kwak-etal-2025-toolhaystack}.

Our work addresses these open challenges by implementing a tool-augmentation approach for agentic gameplay that maintains a structured belief state and injects a corresponding possibility matrix into agent prompts. To this end, we evaluate whether offloading extended-memory belief tracking can improve the deductive reasoning capabilities of autonomous agents, without requiring them to learn when or how to call tools effectively.
\section{Experimental Design}
\label{sec:expdesign}

\subsection{Clue Game Environment}
The murder mystery, board game Clue was adapted into a text-based environment by \cite{ansell2026how}, in which the rules, turn-taking, and game state transitions are conducted through the inputs (prompts) and outputs (generated reasoning logs) of six autonomous agent players. Following the original board game setup, 21 cards (six suspects, six weapons, and nine rooms) are shuffled, and one card from each category is randomly selected and placed in an envelope, forming the hidden solution unknown to all players. The remaining cards are dealt evenly, with each player receiving a private hand of three cards. 

The objective of the game is to correctly deduce the hidden combination of suspect, weapon, and room by making a final accusation. Players take turns making suggestions (i.e., strategic, publicly stated hypotheses) about the solution. If another player holds one or more of the suggested cards, they must privately reveal one of those cards to the suggesting player, with the option to choose which card to show when multiple apply. While other players observe that a card was revealed, they do not see its content. A player will only see at most one card during their turn.

This game interaction structure encourages strategic reasoning and actions, as each turn reveals only partial information; suggestions are publicly expressed through dialogue, while the resulting card reveals remain private to the suggesting player (who may also hold cards included in their own suggestion to potentially mislead other players). In the text-based setting, agents are prompted to verbalize their reasoning at each turn, and their generated reasoning is included in each turn prompt so they can process their previous turns.

In the board game, play ends when the first player makes a correct accusation; however, in our implementation, after a player has won the remaining agentic players are all given a chance to arrive at the final solution and the cards remain hidden until the end. Each game is limited to a maximum of 20 rounds, but ends sooner if all players have made final accusations in less rounds.

\begin{figure*}[ht!]
    \centering
    \includegraphics[width=\textwidth]{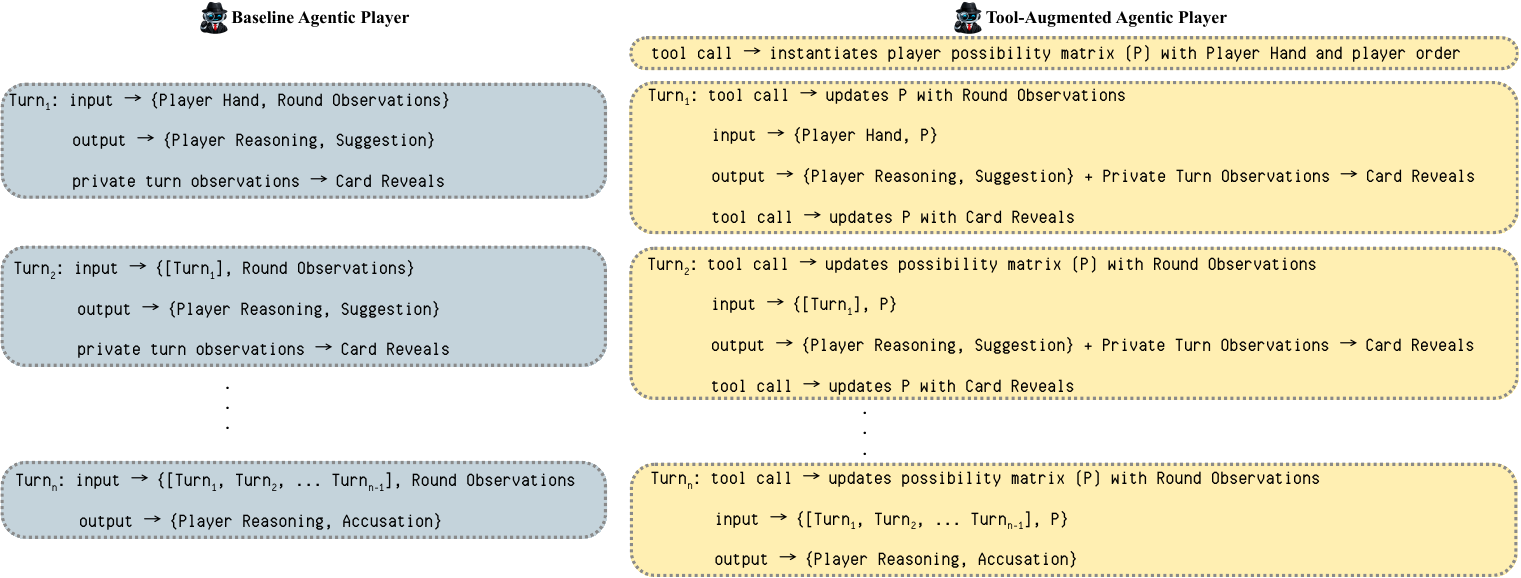}
    \caption{Comparison between the baseline agentic player's turn and the tool-augmented agentic player's turn.}
    \label{fig:toolaug_comparison}
\end{figure*}

\subsection{LLM Agents}
We instantiate agentic players using two language models: OpenAI’s GPT-4o-mini \cite{openai2024gpt4ocard} and Google’s Gemini-2.5-Flash \cite{comanici2025gemini}. Ansell and Toney~\shortcite{ansell2026how} selected these models to balance efficiency and performance, as they are lightweight and cost-effective while still achieving strong reasoning performance. Each agent is initialized with default parameters and receives prompts with identical structure and content across all players. We do not assign distinct personas or role-specific behaviors; all agents operate under the same configuration to ensure consistency and comparability of behavior across model families.

\subsection{Possibility Matrix Tool}

To support belief state tracking beyond verbose agentic reasoning logs, we design a tool that maintains a structured possibility matrix for each agentic player. The tool has three components it maintains for robust player support: \texttt{envelope\_candidates}, \texttt{known\_cards\_by\_player}, and \texttt{game\_matrix}. The tool output represents each agent’s current belief state over all 21 game cards across every possible holder (i.e., all six players and the hidden solution envelope). Each cell in the \texttt{game\_matrix} stores one of three values: YES, NO, or MAYBE, indicating whether a given card is known to be held by a particular player (or the envelope), known to not be held, or remains uncertain.

A player's possibility matrix is instantiated before its first turn action, recording the corresponding hand cards in the \texttt{known\_cards\_by\_player} component and as ``YES'' in the \texttt{game\_matrix}; the remaining cards are populated with ``MAYBE'', for example the \texttt{GPT4o\_MINI\_3} player's \texttt{known\_cards\_by\_player}: 

\begin{verbatim}
    "GPT4o_MINI_3": [
        "Miss Scarlet",
        "Mrs. Peacock",
        "Rope"
        ]
\end{verbatim}

and a component of \texttt{game\_matrix}:
\begin{verbatim}

    "Rope": {
      "ENVELOPE": "NO",
      "GEMINI_FLASH_1": "NO",
      "GEMINI_FLASH_2": "NO",
      "GEMINI_FLASH_3": "NO",
      "GPT4o_MINI_1": "NO",
      "GPT4o_MINI_2": "NO",
      "GPT4o_MINI_3": "YES"
    }
\end{verbatim}

The matrix is then updated incrementally over the course of play based on both private observations and public game events (generated reasoning output and card reveals). Private observations include cards directly revealed to the agent, while public updates are derived from suggestions and disproof behavior. For example, if a player passes on disproving a suggestion, the matrix records that the player does not hold any of the suggested cards (i.e., the matrix records ``NO'' for that player on all three cards). If a player disproves a suggestion but the exact card is not observed, the matrix stores a constraint indicating that the player must hold at least one of the suggested cards. If no player disproves a suggestion, the matrix rules out all non-suggesting players as holders of the suggested cards.

The final reasoning support mechanism of the possibility matrix tool is its evaluation of the remaining envelope candidates (possible solution candidates). If there is exactly one candidate left for each card category (suspect, weapon, and room) in the \texttt{envelope\_candidate} component, the tool includes an \texttt{accusation\_ready} cell that records the solution. The cell remains null until the candidate requirement is satisfied. Thus, agentic players receive explicit information about whether their accumulated knowledge supports making a final accusation.

At each turn, the tool is called and its output is injected into the current agentic player's turn prompt
with the prefix ``Use this matrix state as the authoritative belief state for this turn''. Then the tool is called after the turn actions have ended to update with any cards revealed; this process (and comparison to the baseline implementation) is shown in Figure \ref{fig:toolaug_comparison}. The tool output summary includes current envelope candidates, known card assignments for other players, unresolved disproof constraints, and an indication of whether the agent has sufficient information to make an accusation. In this way, the possibility matrix offloads extended-memory deductive reasoning into a structured intermediate representation. Rather than requiring agents to reconstruct and maintain belief states solely from natural language reasoning history, the tool provides a clean and updated summary of the evolving game state. As a result, agents are able to reason over explicit deductive constraints at each turn.

\subsection{Game Implementation}
We run three versions of the Clue game: \baseline{} (all baseline players), \matrixv{} (all tool-augmented players), and \mixed{} (both baseline and tool-augmented players). Each game has six agentic players (3 per model family), with the \mixed{} game having one tool-augmented player and two baseline players per model family. We run each game version six times, shuffling the player order. Across the 18 games, we record all elements of gameplay (game log, game state, player's hands, player's cards seen, player's prompts, and player's reasoning) for analysis. 

To support answering our three research questions, we evaluate gameplay performance using four metrics: (1) accusation accuracy, (2) deduction quality, (3) knowledge accumulation, and (4) final accusation delay. Accusation accuracy represents the fraction of cards the agentic player correctly used in their accusation (3/3 being perfect). Deduction quality tracks the ratio of correct and incorrect deductions made by players. We label deductions programmatically by evaluating each inferred card assignment against the ground-truth game state, including all player hands and the envelope contents. A deduction is assigned the correct label if it matches the true card assignment and incorrect otherwise. For example, if a player deduces that \texttt{player\_x} holds the Rope card when \texttt{player\_x} does not, that deduction is classified as incorrect. Knowledge accumulation represents the number of cards learned through play per round. Final accusation delay counts how many individual player turns took place from the first suggestion made that had no disproofs (and was the game solution) to the first accusation.

\section{Results}

We summarize the results of the 18 game runs in Table~\ref{tab:summary_conditions}, providing details on agentic player performance across game versions and model families. We report outcome metrics (wins, mean finishing rank, and accusation accuracy) alongside reasoning metrics (mean correct and incorrect deductions per player-game). 

\begin{table*}[ht!]
\centering
\caption{Performance summary across 6 games per condition (36 player-game observations per condition).
\textbf{Outcome}: Games won (out of 6), mean finishing position (Rank; 1\,=\,best, 6\,=\,worst), and normalised accusation accuracy (Acc.; 0--1). \textbf{Reasoning}: Average correct and incorrect deductions per game.}
\label{tab:summary_conditions}
\small
\setlength{\tabcolsep}{5pt}
\begin{tabular}{llcccccc}
\toprule
& & \multicolumn{3}{c}{Outcome}
& \multicolumn{2}{c}{Reasoning} \\
\cmidrule(lr){3-5} \cmidrule(lr){6-7}
Version & Model
& Wins
& Rank
& Acc.
& Ded.~Correct
& Ded.~Incorrect \\
\midrule
\multirow{2}{*}{\baseline{}}
  & GPT-4o-mini      & 4/6 & 3.33 & 0.24 &  9.33 & 2.39 \\
  & Gemini-2.5-Flash & 2/6 & 3.67 & 0.30 &  8.00 & 2.39 \\
\midrule
\multirow{2}{*}{\matrixv{}}
  & GPT-4o-mini      & 3/6 & 3.39 & 0.96 &  6.83 & 1.67 \\
  & Gemini-2.5-Flash & 3/6 & 3.61 & 1.00 &  4.67 & 0.11 \\
\midrule
\multirow{4}{*}{\mixed{}}
  & GPT-4o-mini (baseline) & 4/6 & 3.25 & 0.89 & 7.42 & 1.92 \\
  & GPT-4o-mini (matrix)   & 2/6 & 3.17 & 0.78 & 6.00 & 2.17 \\
  & Gemini-2.5-Flash (baseline) & 0/6 & 4.08 & 0.92 & 7.00 & 1.83 \\
  & Gemini-2.5-Flash (matrix)   & 0/6 & 3.17 & 1.00 & 6.67 & 0.17 \\
\bottomrule
\end{tabular}
\end{table*}

We find that GPT-4o-mini players win the majority of games (13/18) over Gemini-2.5-Flash players. However, all tool-augmented Gemini-2.5-Flash players achieve perfect accusation accuracy, outperforming GPT-4o-mini players.

\subsection{Accusation Accuracy}
\label{sec:accuracy}

\begin{figure*}
    \centering
    \includegraphics[width=\textwidth]{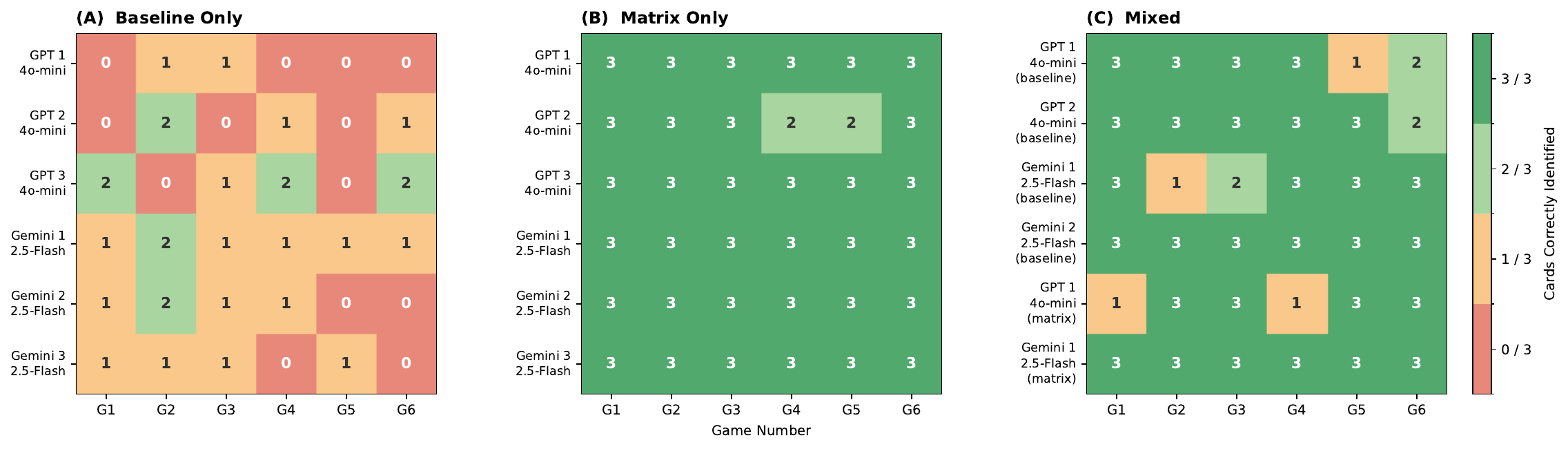}
    \caption{Per-game accusation accuracy for each player, where each cell represents how many solution cards were correctly identified.}
    \label{fig:accuracy_heatmap}
\end{figure*}
We display player-level game accuracy findings in 
Figure~\ref{fig:accuracy_heatmap}. In the \baseline{} game, accuracy was consistently low and no player was able to reach the final solution. Less than half of players (47\%) identified only one card correctly, while 36\% identified none. In contrast, in the \matrixv{} games 94.5\% of all players identified the correct solution, with only two instances of 2/3 accuracy outcomes across the six game observations (from GPT-4o-mini). Gemini-2.5-Flash achieved perfect accuracy once the matrix tool was introduced.

In the \mixed{} run, both player types substantially outperformed the \baseline{} players. Agentic players without tool-augmentation still achieved a mean accuracy of 2.71/3, compared to 0.81/3 in the \baseline{} runs. Tool-augmented agentic players in \mixed{} games achieved a mean accusation value of 2.67/3, comparable to 2.94 from the \matrixv{} game version.

\subsection{Deduction Quality}

\begin{figure}[t]
    \centering
    \includegraphics[width=\columnwidth]{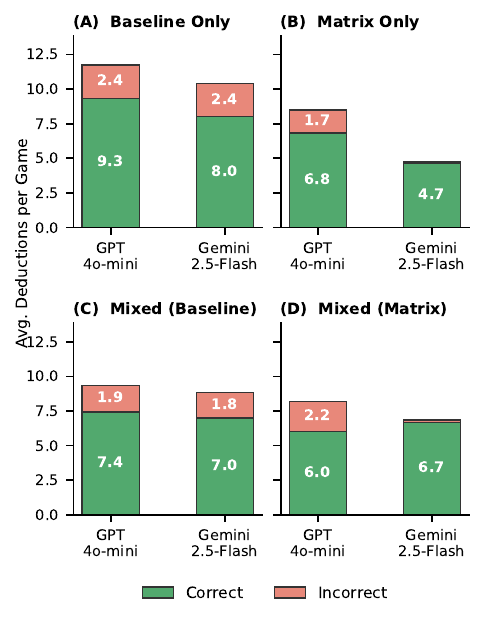}
    \caption{Mean correct (green) and incorrect (red) deductions by player, model family and game version. The top row shows the two homogeneous conditions, Baseline Only (A) and Matrix Only (B), while the bottom row separates Mixed-condition agents by their prompt type: Baseline-prompted (C) and Matrix-prompted (D).}
    \label{fig:deduction_quality}
\end{figure}

Parsing the player logs, we programmatically evaluate if deductions made are viable (i.e., did a player make a deduction that is contradictory to the actions and observations in the game) and denote viable deductions as correct. Figure~\ref{fig:deduction_quality} shows the mean correct and incorrect deductions per player broken down by game version and model family. In the \baseline{} version (A), both models produced the highest raw counts of correct deductions (GPT-4o-mini: 9.3, Gemini-2.5-Flash: 8.0), but also the highest incorrect deduction counts (2.4 each).

The \matrixv{} game reduced the number of incorrect deductions for both models, most noticeably for Gemini-2.5-Flash, which dropped to 0.1 incorrect deductions per game while making 4.7 correct deductions. GPT-4o-mini also improved, reaching 6.8 correct and 1.7 incorrect deductions. The overall reduction in both correct and incorrect counts relative to Baseline is consistent with games resolving faster when a tool-augmented agent is playing, leaving fewer rounds for inference.

In the \mixed{} condition, the deduction quality pattern differs by player type. Baseline players (C) produced correct deduction rates slightly lower than the \baseline{} game. Matrix players (D) showed similar profiles to the (B) \matrixv{} game, with GPT-4o-mini performing slightly worse, and Gemini Flash performing slightly better.

\subsection{Knowledge Accumulation}

\begin{figure*}[t]
    \centering
    \includegraphics[width=\textwidth]{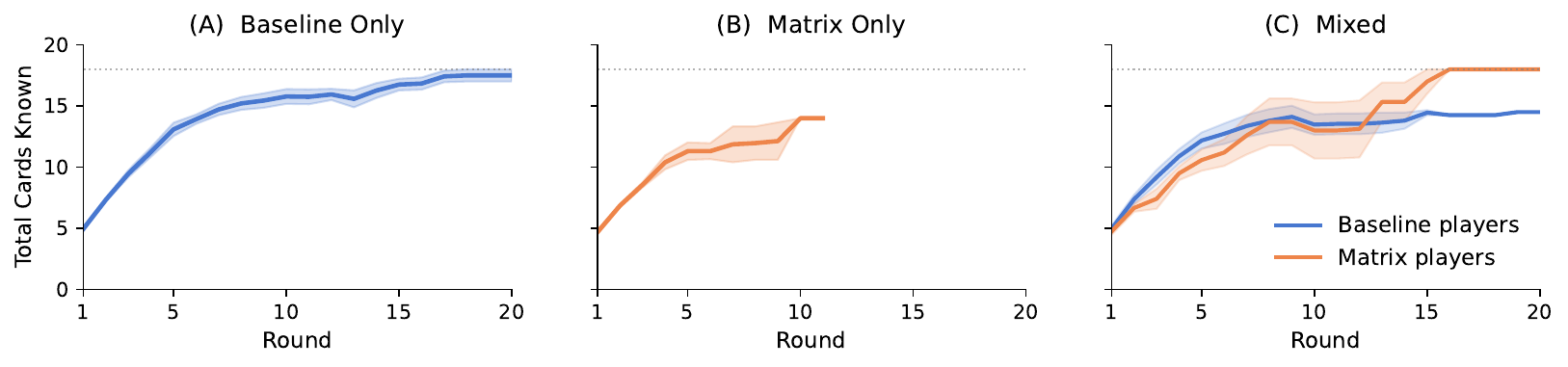}
    \caption{Mean cards learned through play per round, averaged across all players within each simulation. Shaded bands show the standard error mean across games. The dotted line at 18 marks the the ceiling (18 non-solution cards minus the 3 dealt to each player.) }
    \label{fig:knowledge_curve}
\end{figure*}

We compute the number of new cards discovered per game round by player to evaluate knowledge accumulation. 
Figure~\ref{fig:knowledge_curve} shows total cards known per round (player's hand plus cards seen through play) averaged across games, with a maximum of 18 known cards. Across the three game versions, we find that tool-augmented players accumulate knowledge more slowly per round, but translate what they learn into correct accusations more efficiently, as seen in Section \ref{sec:accuracy}, while Baseline players rely on longer games with extended play to build a near complete picture before solving. 

Baseline players (A) accumulate knowledge rapidly in the first five rounds and approach the knowledge ceiling by round 17--18, utilizing the full 20-round game length to observe suggestions and build up a complete picture. In contrast, tool-augmented players (B) end their games much earlier (maximum of 11 rounds observed), truncating the knowledge curve around 14 cards. These players never reach the same total knowledge as the Baseline players, but they solve the game far more accurately. 
The \mixed{} game (C) shows the contrast between the two methods side by side. The baseline players gather information slightly quicker, but plateau around 14--15 cards, while the Matrix players diverge sharply after round 10, reaching the ceiling of 18 cards by round 15.


\subsection{Solution Suggested to Final Accusation}

By measuring the number of turns between a correct suggestion (i.e., the suggestion is the game solution) and the subsequent correct accusation, we estimate how effectively players leverage accumulated information to form strong deductions and reach a final decision. Additionally, this analysis provides insights into how ``risky'' players are based on their game state beliefs. 

\begin{figure*}[t]
    \centering
    \includegraphics[width=\textwidth]{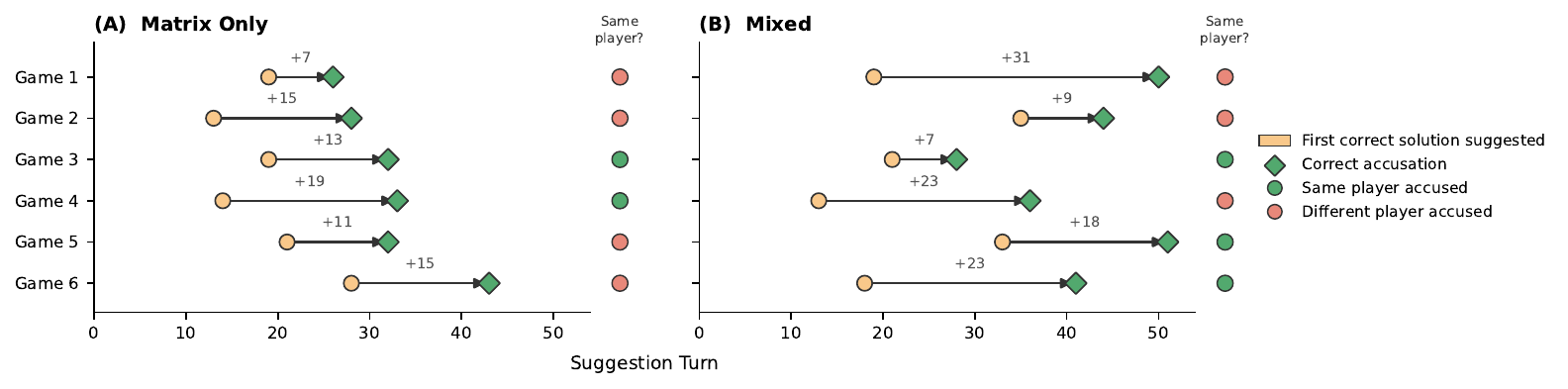}
    \caption{Per-game solution timing in the Matrix~(A) and Mixed~(B) conditions. Each row is one game. The orange circle marks the first turn on which the complete solution was suggested and went undisproved; the green diamond marks the turn of the correct accusation. Gap labels (+\textit{N}) show the number of turns (6 turns per round). The side column indicates whether the same player who first identified the solution also made the final accusation (green) or a different player did (red).}
    \label{fig:solution_timing}
\end{figure*}

Figure~\ref{fig:solution_timing} plots the turn where the correct solution was first publicly confirmed (all three solution cards suggested, no player able to disprove) and the turn where the final correct accusation ended the game.

In the \matrixv{} game, the first confirmed suggestion occurred between turns 18 and 23, and the accusation followed 7--19 turns later. Only 2 of the 6 games were won by the player who first suggested the solution. In the \mixed{} game, the first confirmed suggestion occurred between turns 13 and 35, and the gap to accusation was 7--31. The first suggester won 3 out of 6 games. 

To understand the delay at the player level from knowing the solution to final accusation, we examined when each tool-augmented player first received \texttt{accusation\_ready} from the possibility matrix, indicating that there was only one candidate per card category remaining. In the \matrixv{} game, 33 out of 36 players had \texttt{accusation\_ready} set before the game ended. The mean gap from this point to the game-ending accusation was 11.1 turns for GPT-4o-mini and 8.2 turns for Gemini-2.5-Flash. In the \mixed{} game version (only tool-augmented players received the signal) their mean gaps were 18.0 turns (GPT-4o-mini) and 15.3 (Gemini-2.5-Flash). Inspection of the player-turns where the \texttt{accusation\_ready} was set but no accusation was made shows a recurring pattern: agents do not exhibit risky play. All agentic players gather more information despite knowing with certainty what the final solution was.

\section{Discussion}


Here, we address the answers to our research questions drawing evidence from our presented results.

\paragraph{(RQ1) To what extent does externalized belief state tracking via tool augmentation improve agent reasoning and gameplay performance? }
Comparing the accusation accuracies across game versions, we find that tool augmentation has a clear effect: games with tool-augmented players consistently achieve near-perfect accuracy. This observed gameplay improvement cannot be attributed to increased information exposure (highlighted in Figure \ref{fig:knowledge_curve}). Baseline agents approached the knowledge ceiling of cards by the final rounds and produced the highest raw deduction counts of any game version; however, they failed to convert this evidence into correct accusations. Tool-augmented agents accumulated substantially less total knowledge, but they made accusations with near perfect accuracy. 

The response of the two models to the possibility matrix tool differs slightly. Gemini-2.5-Flash reduces incorrect deductions to near zero and achieves perfect accusation accuracy, suggesting that it treats the matrix as a hard constraint. GPT-4o-mini retains a similar rate of incorrect deductions across both conditions, indicating that structured output does not fully remove unsupported reasoning for all models. Despite the differences in deduction precision, both models achieve comparable win counts, suggesting that the \matrixv{} game outcomes are more balanced.

\paragraph{(RQ2) How does the presence of tool-augmented agents influence the behavior and performance of non-tool agents in mixed-agent environments? }

Baseline-prompted players in the \mixed{} games achieved substantially higher accusation accuracy than in the \baseline{} games, despite no change to their prompt. Their deduction profiles across the two experiments are fairly similar, and so their accusation accuracy improvement reflects a possible game environment effect: tool-augmented players producing more informative suggestions raises the quality of shared evidence available to all players regardless of prompt type.

More specifically, this finding suggests that the benefits of structured belief states are not confined to the agents that use it directly. Even a minority of tool-augmented agents is sufficient to improve group-level accusation accuracy close to that of a fully tool-augmented player game. At the same time, the win distribution in the \mixed{} games reveals an asymmetry among the model families, as all wins were taken by GPT-4o-mini players regardless of prompt type, while Gemini-2.5-Flash players won none in either role. This observed pattern suggests that win outcomes in \mixed{} games may be more affected by model behavior (e.g., willingness to accuse) than by whether or not a player used the possibility matrix tool. 


\paragraph{(RQ3) How does structured reasoning support affect decision-making, particularly with respect to risk-taking and the timing of final decisions?}

In our experiments, we identified a notable turn gap between the solution knowledge and accusation. The \texttt{accusation\_accuracy} analysis available with the tool-augmented players suggests that agents were not strictly using the structured belief representation for their action selection. This observed behavior indicates that agentic players retain autonomy in their decision-making, as we did not require that they accuse when they received the \texttt{accusation\_accuracy} signal, only that they use the information in the possibility matrix to make their turn. Adjusting the timing of final accusations would likely require a prompt-based intervention; these results are consistent with prior work highlighting tool-use challenges of autonomous agents. 

Our hypothesized bottleneck in unstructured game play was the ability to maintain a logically consistent belief state over sequential observations, as Ansell and Toney~\shortcite{ansell2026how} found that text-based fine-tuning on a related task worsened agentic play. We found that our tool-augmented approach enables the models to access a structured representation of the game state, while preventing the accumulation of unsupported inferences that are characteristic of the Baseline players' reasoning. This effect is highlighted in the  reduction of incorrect deductions across both models with the introduction of the possibility matrix tool.

In summary, our findings suggest that tool augmentation targeting the offloading of extended-memory reasoning and belief tracking leads to substantial improvements in tasks requiring sustained deductive inference over sequential, long interactions. However, tool-augmentation (as we implemented it) does not necessarily directly affect the autonomy of agents in their decision-making. 

\section{Limitations and Future Work}
While our findings provide evidence that structured belief tracking can support deductive reasoning, we outline several limitations as follows:

First, our proposed possibility matrix is a task-specific representation designed around the deductive structure of the game \textit{Clue}. The matrix explicitly encodes card ownership constraints, maintains a structured belief state, and provides an \texttt{accusation\_ready} signal when only a single solution remains. As a result, the tool is designed for a specific environment as opposed to a generalizable solution. Future work should investigate whether similar representations generalize beyond \textit{Clue} to other reasoning domains, including tasks requiring inductive or abductive reasoning, as well as other deductive environments with different state structures and constraints.


Second, the experimental evaluation is limited in scale. We evaluate two model families (Gemini-2.5-Flash and GPT-4o-mini) across 18 total games, with six games per experimental condition. Although this setup produces consistent qualitative trends, the number of models and independent game runs is small-scale for a stochastic, multi-agent environment. Our experiments provide preliminary results suggesting that external belief tracking tools are a promising approach for improved agentic deductive reasoning. Future work should conduct larger-scale evaluations across additional models and game simulations to assess the robustness and generality of these results. In particular, comparisons involving larger reasoning-oriented models may help clarify whether the observed benefits arise primarily from compensating for limitations in lightweight models or represent a broader advantage of structured belief tracking.




Lastly, our baseline comparisons focus on agents that reason from natural-language game histories. While this establishes a clear contrast between unstructured and structured state representations, it does not isolate which aspects of the possibility matrix are responsible for the observed improvements. For example, deduction and gameplay improvement may arise from reduced context length, explicit constraint representation, improved memory retention, or some combination of these factors. Future work could compare against intermediate baselines, such as summarized histories, model-generated tables, or alternative memory-augmentation strategies.


\section{Conclusion}

Achieving strong performance in extended-memory deductive reasoning tasks remains challenging for autonomous agents. In this work, we investigate whether offloading belief state tracking (previously maintained through natural language logs) to an external tool can improve agent performance. We evaluate this approach within a text-based implementation of the board game Clue, a dynamic multi-agent environment in which outcomes depend on the interactions, strategies, and information exchange between players, rather than static question-answering tasks with fixed ground truth (e.g., “What is the capital of France”). 

Our results show that tool augmentation substantially improves gameplay performance, achieving near-perfect accuracy for tool-augmented agents while also benefiting baseline agents in mixed settings. More broadly, these results suggest that when belief states are presented to an agent in a more succinct and structured representation (over natural language reasoning logs) lightweight agents can perform near-optimally in complex, interactive settings.

\appendix

\bibliographystyle{named}
\bibliography{8_bib}

\end{document}